%% file: 0-main.tex
\documentclass[letterpaper, 10 pt, conference]{ieeeconf}  % Comment this line out if you need a4paper

\IEEEoverridecommandlockouts                              % This command is only needed if 
\usepackage{multicol}
\usepackage[bookmarks=true]{hyperref}

\usepackage{mathtools}
\usepackage{amsfonts}  % for \mathbb
\usepackage{graphics}
\usepackage[pdftex]{graphicx}
\usepackage{wrapfig}
\usepackage{color}
\usepackage{dsfont}
\usepackage{algorithmicx}
\usepackage[ruled]{algorithm}
\usepackage{algpseudocode}
\usepackage{amssymb}

\usepackage{caption}
\usepackage{wrapfig}

\algnewcommand\algorithmicforeach{\textbf{for each}}
\algdef{S}[FOR]{ForEach}[1]{\algorithmicforeach\ #1\ \algorithmicdo}

\DeclarePairedDelimiterX{\infdivx}[2]{(}{)}{%
  #1\;\delimsize\|\;#2%
}

\usepackage{expl3}
\ExplSyntaxOn
\newcommand\latinabbrev[1]{
  \peek_meaning:NTF . {% Same as \@ifnextchar
    #1\@}%
  { \peek_catcode:NTF a {% Check whether next char has same catcode as \'a, i.e., is a letter
      #1.\@ }%
    {#1.\@}}}
\ExplSyntaxOff
\def\eg{\latinabbrev{e.g}}

\def\etc{\latinabbrev{etc}}
\def\ie{\latinabbrev{i.e}}

\title{
\LARGE \bf
Planning to Practice: Efficient Online Fine-Tuning \\by Composing Goals in Latent Space
}

\author{
Kuan Fang$^*$, Patrick Yin$^*$, Ashvin Nair, Sergey Levine\\
University of California, Berkeley
\thanks{$^{*}$Authors contributed equally to this work.}
}

\begin{document}

\maketitle
\thispagestyle{empty}
\pagestyle{empty}

\begin{abstract}
General-purpose robots require diverse repertoires of behaviors to complete challenging tasks in real-world unstructured environments. To address this issue, goal-conditioned reinforcement learning aims to acquire policies that can reach configurable goals for a wide range of tasks on command. However, such goal-conditioned policies are notoriously difficult and time-consuming to train from scratch. In this paper, we propose Planning to Practice (PTP), a method that makes it practical to train goal-conditioned policies for long-horizon tasks that require multiple distinct types of interactions to solve. Our approach is based on two key ideas. First, we decompose the goal-reaching problem hierarchically, with a high-level planner that sets intermediate subgoals using conditional subgoal generators in the latent space for a low-level model-free policy. Second, we propose a hybrid approach which first pre-trains both the conditional subgoal generator and the policy on previously collected data through offline reinforcement learning, and then fine-tunes the policy via online exploration. This fine-tuning process is itself facilitated by the planned subgoals, which breaks down the original target task into short-horizon goal-reaching tasks that are significantly easier to learn. We conduct experiments in both the simulation and real world, in which the policy is pre-trained on demonstrations of short primitive behaviors and fine-tuned for temporally extended tasks that are unseen in the offline data. Our experimental results show that PTP can generate feasible sequences of subgoals that enable the policy to efficiently solve the target tasks.  
\footnote{Supplementary video: \href{https://sites.google.com/view/planning-to-practice}{sites.google.com/view/planning-to-practice}}

\end{abstract}

\input{1-introduction}
\input{2-related-work}
\input{3-background}
\input{4-method}
\input{5-experiments}
\input{6-conclusion}

% \newpage
% \vspace{-1.5mm}
{\small
\bibliographystyle{bibtex/IEEEtran}
\bibliography{bibtex/references}
}

\newpage
\clearpage

\end{document}

%% file: 1-introduction.tex
\section{Introduction}

% %% What is the problem?
% Utilizing large datasets for robotics -> GCRL from offline data
Developing controllers that can solve a range of tasks and generalize broadly in real-world settings is a long-standing challenge in robotics.
Such generalization in other domains such as computer vision and natural language processing has been attributed to training large machine learning models on massive datasets~\cite{krizhevsky2012imagenet}.
Consequently, one promising approach to robustly handle a wide variety of potential situations that a robot might encounter is to train on a large dataset of robot behaviors.
Prior work in robotics has demonstrated effectively generalization from training on large datasets for individual tasks, such as grasping~\cite{levine2017grasping, yu2021conservative}.
However, a general-purpose robot should be able to perform a wide range of skills, and should also be \textit{taskable}.
That is, it must be able to complete a specific task when specified by a human, including temporally extended tasks that require sequencing many skills together to complete.
This topic has been studied in prior work on goal-conditioned reinforcement learning, where a robot aims to perform a task given a desired end state~\cite{Khazatsky2021WhatCI, chebotar2021actionable, kalashnikov2021mtopt}.
What remains to make these methods widely applicable for real-world robotics?

\begin{figure}[t!]
    \centering
    \includegraphics[width=.4\textwidth]{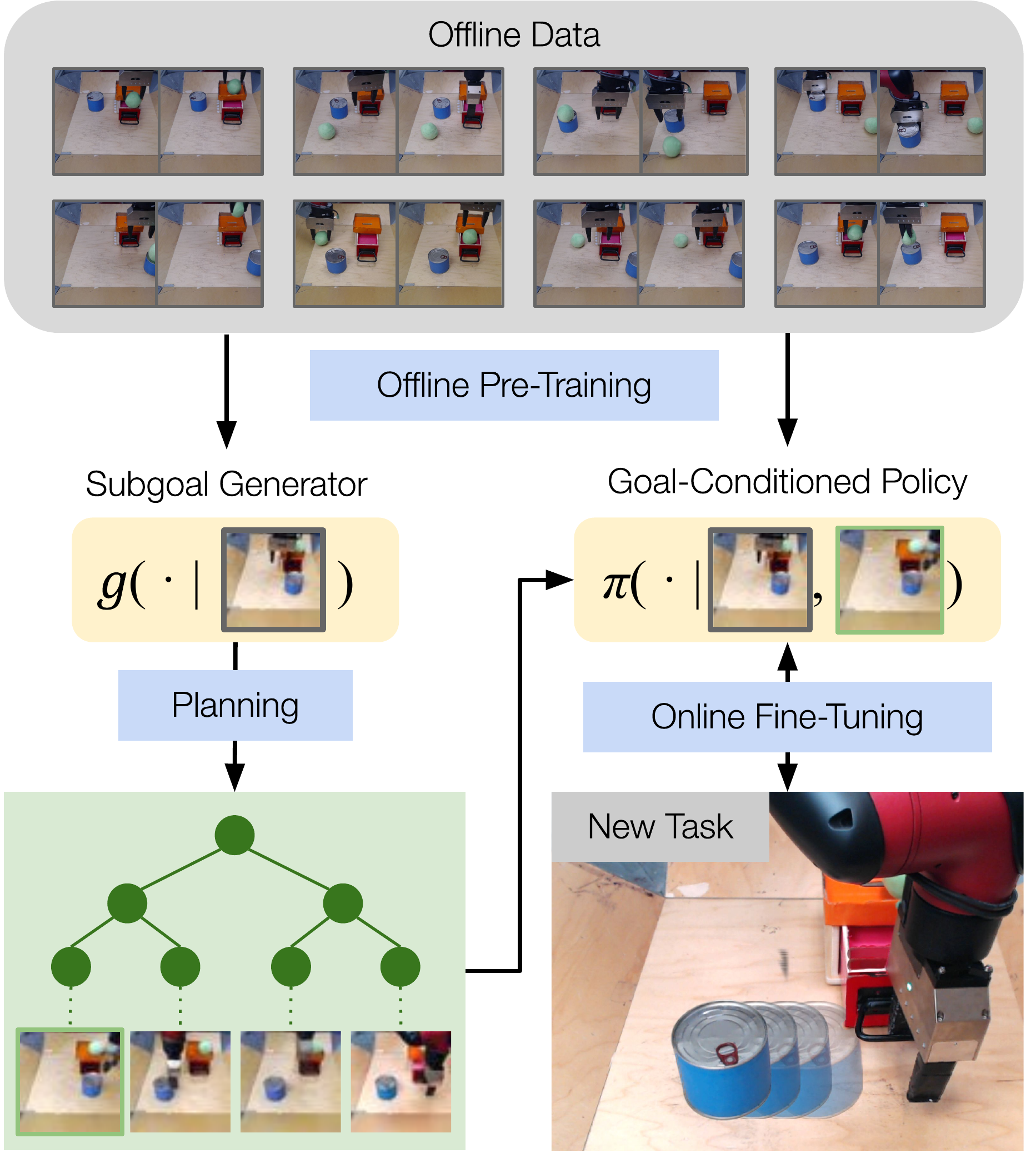}
    % \includegraphics[width=.4\textwidth]{example-image-a}
    % \vspace{-6mm}
    \caption{Our method, Plan to Practice (PTP), solves long-horizon goal-conditioned tasks by combining planning and fine-tuning.
    We begin with an offline dataset containing a variety of behaviors, and train a subgoal generator and goal-conditioned policy on this data. Then, to learn more complex multi-stage tasks, we optimize over subgoals using the subgoal generator, which corresponds to a planning procedure over (visual) subgoals, and fine-tune the policy with online RL by practicing these subgoals. This enables the robot to solve multi-stage tasks directly from images.}
    \label{fig:intro}
    \vspace{-5mm}
\end{figure}

While goal-conditioned policies can be trained effectively for relatively short-horizon tasks, temporally extended multi-stage can pose a significant challenge for current methods. These tasks present a major exploration challenge during online learning, and a major challenge for credit assignment during offline learning.In this paper, we aim to address these challenges by combining two ideas. The first is that long-horizon goal-reaching tasks can be decomposed into shorter-horizon tasks consisting of subgoals. The second is that these subgoals can be used to \emph{fine-tune} a goal-conditioned policy online, even if its performance from offline data is poor.The first idea enables us to address the exploration challenge, by automatically generating intermediate subgoals that can be ``practiced'' on the way to a longer-horizon final goal. In the framework of goal-conditioned RL, solving long-horizon tasks can be reduced to the problem of optimization over a sequence of subgoals for the goal-conditioned policy, and this optimization over subgoals can be regarded as a kind of high-level planning, where the optimizer selects waypoints for achieving a distant goal. The high-level planner itself can use a learned high-level model.

However, if we rely entirely on offline data, credit assignment challenges make it difficult to perform longer-horizon tasks even with subgoal planning. Even if the offline RL policy performs well on each individual skill, there may be errors from stitching skills together because the initial states of each stage diverge from the offline data when they are composed together. In practice, this leads to poor performance when using only offline training. Therefore, the second key idea in our work is to utilize subgoal planning not merely to \emph{perform} a multi-stage task, but also to make it possible to \emph{practice} that task to finetune it online. While online training for temporally extended tasks is ordinarily difficult, by addressing the exploration challenge with subgoal planning, we make it possible for the robot to practice a series of relatively short-horizon tasks, which makes this kind of finetuning feasible. Thus, the planner acts both as a higher level policy when performing the task, and as a scaffolding curriculum for finetuning the lower-level goal-conditioned policy
By collecting data actively in a specific environment, we can directly experience the distribution shift and can use reinforcement learning to improve performance under this shift.

To this end, we propose Planning to Practice (PTP), an approach that efficiently trains a goal-conditioned policy to solve
multi-step tasks by setting subgoals to exploit the compositional structure of the offline data.
An outline is shown in Fig.~\ref{fig:intro}. Our approach is based around a planner that composes generated subgoals
to guide the goal-conditioned policy during an online fine-tuning phase.
To propose diverse and reachable subgoals to form the candidate plans, we design a conditional subgoal generator based on conditional variational autoencoder (CVAE)~\cite{sohn2015cvae}.
Through training on the offline dataset, the conditional subgoal generator captures the distribution of reachable subgoals from a given state and generates sequences of subgoals from the learned latent space in a recursive manner.
Our subgoal planning algorithm hierarchically searches for subgoals in a coarse-to-fine manner using multiple conditional subgoal generators that are trained to generate goals at different temporal resolutions.
Both the goal-conditioned policy and the conditional subgoal generators are pre-trained on the offline data, and the policy is fine-tuned on the novel target task. 

% Summary of contributions
Our main contribution is a system for learning to solve long-horizon goal-reaching tasks by fine-tuning the goal-conditioned policy with subgoal planning in a learned latent space. 
We evaluate our approach on multi-stage robotic manipulation tasks with raw image observations and image goals in both simulation and the real world.
After being pre-trained on short demonstrations of primitive interactions, our approach is able to find feasible subgoal sequences as plans for unseen final goals by recursively generating subgoals with the learned conditional subgoal generators. By comparing our approach with both model-free methods and prior approaches that optimize over subgoals, we demonstrate that the produced plans significantly improve the learning efficiency and the resultant success rates during the online fine-tuning.

%% file: 2-related-work.tex
\section{Related Work}

We propose to use a combination of optimization-based planning and fine-tuning with goal-conditioned reinforcement learning from prior data in order to allow robots to learn temporally extended skills. In this section, we cover prior methods in offline RL, planning, goal-conditioned RL, and how they relate to our method.

\textbf{Learning from prior data.}
Offline reinforcement learning methods learn from prior data~\cite{lange2012batch, fujimoto2019off, kumar2019stabilizing, zhang2021brac, kumar2020conservative, fujimoto2021minimalist,singh2020cog}, and can also finetune through online interaction~\cite{nair2020awac, villaflor2020finetuning, lu2021awopt, Khazatsky2021WhatCI, lee2021finetuning, meng2021starcraft}. Such methods have been used in a variety of robotic settings~\cite{kalashnikov2018scalable,cabi2019scaling,kalashnikov2021mtopt,lu2021awopt}. Our focus is not on introducing new offline RL methods. Rather, our work shows that planning over subgoals for a goal-conditioned policy that is pretrained offline can enable finetuning for temporally extended skills that would otherwise be very difficult to learn.

\textbf{Goal-conditioned reinforcement learning.} The aim of goal-conditioned reinforcement learning (GCRL) is to control the agent to efficiently reach specified goal states~\cite{Kaelbling1993LearningTA, Schaul2015UniversalVF, Eysenbach2021CLearningLT}. Compared to policies that are trained to solve a fixed task, the same goal-conditioned policy can perform a variety of tasks when it is commanded with different goals. Such flexibility allows GCRL to better share knowledge across different tasks and make use of goal relabeling techniques to improve the sample efficiency without meticulous reward engineering~\cite{Andrychowicz2017HindsightER, Pong2020SkewFitSS, Fang2019CurriculumguidedHE, Ding2019GoalconditionedIL, Gupta2019RelayPL, Sun2019PolicyCW, Eysenbach2020RewritingHW, Ghosh2021LearningTR}. Prior has explored various strategies for proposing goals for exploration~\cite{nair2018rig, Nair2019ContextualIG, Khazatsky2021WhatCI, ChaneSane2021GoalConditionedRL}, and studied goal-conditioned RL from offline data~\cite{chebotar2021actionable}. However, such works generally aim to learn short-horizon behaviors, and learning to reach goals that require multiple stages (e.g., several manipulation primitives) is very difficult, as shown in our experiments. Our work aims to extend model-free goal-conditioned RL methods by incorporating elements of planning to enable effective finetuning for multi-stage tasks.

\textbf{Planning.} A wide range of methods have been developed for planning in robotics. At the most abstract level, symbolic task planning searches over discrete logical formulas to accomplish abstract goals~\cite{fikes1971strips}. Motion planning methods solve the geometric problem of reaching a goal configuration with dynamics and collision constraints~\cite{Kavraki1996, koenig2002dstarlite,  karaman2011rrtstar, zucker2013chomp, kalakrishnan2011stomp}. Prior methods have also considered task and motion planning as a combined problem~\cite{srivastava14tamp}. These methods generally assume high-level structured representations of environments and tasks, which can be difficult to actualize in real-world environments. Since in our setting we only have image inputs and not structured scene representations, we focus on methods that can handle raw images for observations and task specification.

\textbf{Combining goal-conditioned RL and planning.}
A number of recent works have sought to integrate concepts from planning with goal-conditioned policies in order to plan sequences of subgoals for longer-horizon tasks~\cite{Nasiriany2019PlanningWG, Eysenbach2019SearchOT, fang2019cavin, Charlesworth2020PlanGANMP, Pertsch2020LongHorizonVP, Sharma2021AutonomousRL, Zhang2021CPlanningAA}. These prior methods either propose subgoals from the set of previously seen states, or directly optimize over subgoals, often by utilizing a latent variable model to obtain a concise representation of image-based states~\cite{nair2018rig,ichter2018learning,nair2019hierarchical,Nasiriany2019PlanningWG,Pertsch2020LongHorizonVP, Khazatsky2021WhatCI, ChaneSane2021GoalConditionedRL}. 
The method we employ is most closely related conceptually to the method proposed by Pertsch et al.~\cite{Pertsch2020LongHorizonVP}, which also employs a hierarchical subgoal optimization, and the method proposed by Nasiriany et al.~\cite{Nasiriany2019PlanningWG}, which also optimizes over sequences of latent vectors from a generative model. Our approach makes a number of low-level improvements, including the use of a conditional generative model~\cite{Nair2019ContextualIG}, which we show leads to significantly better performance. More importantly, our method differs conceptually from these prior works in that our focus is specifically on utilizing subgoal optimization as a way to enable finetuning goal-conditioned policies for longer-horizon tasks. We show that it is in fact this capacity to enable effective finetuning that enables our method to solve more complex multi-stage tasks in our experiments.

%% file: 3-background.tex
\section{Problem Statement}

In this paper, we consider the problem of learning to complete a long-horizon task specified by a goal image.
The robot learns over a variety of initial configurations and goal distributions, which cover a range of behaviors such as opening or closing a drawer, and picking, placing, or pushing an object.
As prior data, the robot has access to an offline dataset of trajectories $\mathcal{D}_\text{offline} = \{\tau_1, \tau_2, \dots, \tau_N\}$ for offline pre-training.
In each trajectory, the robot is controlled by a human tele-operator or a scripted policy to achieve one of the goals the environment affords.
A goal-conditioned policy is pre-trained on this dataset using offline RL algorithms. 

After offline pre-training, the robot isplaced in a particular environment that it has online access to interact in.
Even though the initial configuration of this environment may have been included in the set of training environments, the goal distribution for this environment at test time requires sequencing multiple skills together, which is not present in the offline data.
For instance, as shown in Fig.~\ref{fig:intro}, the robot would need to first slide away the can that blocks the drawer, then reaches the handle of the drawer, and finally opens the drawer. 

Na\"ively running offline RL may not solve the long-horizon test tasks for two reasons.
First, the robot is given a test goal distribution that is long horizon but offline dataset consists of individual skills.
The method needs to somehow compose these individual skills autonomously in order to succeed at goals drawn from the test distribution.
Second, offline RL may not solve the task due to distribution shift.
Distribution shift appears in two forms: distribution shift between transitions in the prior data and transitions obtained by the actively rolling out the policy, and the distribution shift introduced when performing tasks sequentially.
If the robot may actively interact in the new environment to improve its policy, how can the robot further practice and improve its performance?

\section{Preliminaries}
\label{sec:preliminaries}
We consider a goal-conditioned Markov Decision Process (MDP) denoted by a tuple $M = (\mathcal{S},\mathcal{A}, \rho, P, \mathcal{G}, \gamma)$ with state space $\mathcal{S}$, action space $\mathcal{A}$, initial state probability $\rho$, transition probability $P$, a goal space $\mathcal{G}$, and discount factor $\gamma$. In each episode, a desired goal $s_g \in \mathcal{G}$ is sampled for the robot to reach. At each time step $t$, a goal-conditioned policy $\pi(a_t | s_t, s_g)$ selects an action $a_t \in \mathcal{A}$ conditioned on the current state $s_t$ and goal $s_g$. After each step, the robot receives the goal-reaching reward $r_t(s_{t+1}, s_g)$.
The robot aims to reach the goal by maximizing the average cumulative reward $\mathbb{E}[\Sigma_t \gamma^t r_t]$.
% ----------------------------------------------------------------------------------------
Our approach learns a goal-conditioned policy $\pi$ for solving the target task specified by a desired final goal $s_g$. The goal-conditioned policy is pre-trained on a previously collected offline dataset $\mathcal{D}_\text{offline}$ and then fine-tuned to reach $s_g$ by accumulating data into an online replay buffer $\mathcal{D}_\text{online}$. $\mathcal{D}_\text{offline}$ contains diverse short-horizon interactions with objects in the environment. During online fine-tuning, we would like the policy to learn to improve and compose these short-horizon behaviors for multi-stage tasks specified by $s_g$. 

Defining informative goal-reaching rewards and extracting useful state representations from high-dimensional raw observations such as images can be challenging. Following the practice in prior work~\cite{nair2018rig, Khazatsky2021WhatCI}, we pre-train a state encoder $h=\phi(s)$ to extract the latent state representation $h$. By encoding the states and goals to the latent space, we can obtain an informative goal-reaching reward function $r_t = R(h_{t+1}, h_g)$ by computing $h_{t+1} = \phi(s_{t+1})$ and $h_g = \phi(s_g)$. Specifically, $R(h_{t+1}, h_g)$ returns 0 when $||h_{t+1}, h_g || < \epsilon$ and -1 otherwise, where $\epsilon$ is a selected threshold. In addition, we also use $\phi(s_{t+1})$ as the backbone feature extractor in all of our models that take $s$ as an input. For simplicity, we directly use $s$ to denote $h$ in the rest of the paper. The details of the state encoder are explained in Sec.~\ref{sec:implementation_details}.

%% file: 4-method.tex
\section{Planning to Practice}

We propose Planning to Practice (PTP), an approach that efficiently fine-tunes a goal-conditioned policy to solve novel tasks.  
To enable the robot to efficiently learn to solve the target task, we propose to use subgoals to facilitate the online fine-tuning of the goal-conditioned policy. Given the initial state $s_0$ and the goal state $s_g$, we search for a sequence of $K$ subgoals $\hat{s}_{1:K} = \hat{s}_1, ..., \hat{s}_K$ to guide the robot to reach $s_g$. Such subgoals will inform the goal-conditioned policy $\pi(a | s, s_g)$ what is the immediate next step on the path to $s_g$ and provide the policy more dense reward signals compared to directly using the final goal. We choose the sequence of subgoals at the beginning of each episode and feed the first subgoal in the sequence to the goal-conditioned policy. The policy will switch to the next subgoal in the sequence when the current subgoal is reached or the time budget assigned for the current subgoal runs out.

The main challenge is to search for a sequence of subgoals that can lead to the desired final goals while ensuring each subgoal is a valid state that can be reached from the previous subgoal. Particularly when the states correspond to full images, most vectors will not actually represent valid states, and indeed na\"{i}vely optimizing over image pixels may simply result in out-of-distribution inputs that lead to erroneous results when input into the goal-conditioned policy.

As outlined in Fig.~\ref{fig:intro}, we devise a method to effectively propose and select valid subgoal sequences to guide online fine-tuning by means of a generative model. At the heart of our approach is a conditional subgoal generator $g(\cdot | s_0)$ that recursively produces candidate subgoals in a hierarchical manner conditioned on the initial state $s_0$. To find the optimal sequence of subgoals $\hat{s}_{1:K}^*$, we first sample $N$ candidate sequences $\hat{s}_{1:K}^1, ..., \hat{s}_{1:K}^N$ from the state space using the conditional subgoal generator. Then we rank the candidate sequences using a cost function $c$. The sequence that corresponds to the lowest cost will be selected as $\hat{s}_{1:K}^*$ for the goal-conditioned policy. Through this sampling-based planning procedure, we choose the subgoal for guiding the goal-conditioned policy $\pi$ during online fine-tuning. The overall algorithm is summarized in Algorithm~\ref{algo:ptp}. Next we describe the design of each module in details.

% ----------------------------------------------------------------------------------------
\begin{algorithm}[t]
\caption{Planning To Practice (PTP)}
\begin{algorithmic}[1]
\Require set of final goals $\mathcal{G}$, time horizon $T$, offline data $\mathcal{D}_\text{offline}$, number of subgoals $K$.

\State Train $\pi(a | s, s_g)$ and $g(s, z)$ on $\mathcal{D}_\text{offline}$.
\State Initialize the online replay buffer $\mathcal{D}_\text{online} \leftarrow \varnothing$.

\While{not converged}
    \State Reset the environment and observe $s_0$.
    \State Sample $s_g$ from $\mathcal{G}$.
    \State Plan for the subgoals $\hat{s}_{1:K}$.
    
    \State $k \leftarrow 1$
    \For{$t = 1, ..., T$}
        \State Compute the action $a_t \leftarrow \pi(a_t | s_t, \hat{s}_k)$
        \State Observe the state $s_{t+1}$ and the reward $r_t$
        \State $\mathcal{D}_\text{online} \leftarrow \mathcal{D}_\text{online} \cup (s_t, a_t, r_t, s_{t+1})$.
        
        \If{$t \pmod {\Delta t} == 0$ \textbf{or} $|| s_{t+1} - \hat{s}_K || < \epsilon$}
            \State $k \leftarrow \min(k + 1, K)$ 
        \EndIf
    \EndFor
    
    \State Train $\pi$ on batches sampled from $\mathcal{D}_\text{offline}$ and $\mathcal{D}_\text{online}$.
    
\EndWhile

\end{algorithmic}
\label{algo:ptp}
\end{algorithm}

% ----------------------------------------------------------------------------------------
\subsection{Conditional Subgoal Generation}

The effectiveness of our planner relies on the generation of diverse and feasible sequences of subgoals as candidates. Specifically, we would like to generate the candidates by sampling from the distribution of suitable subgoal sequences $p(\hat{s}_1, ..., \hat{s}_K | s_0)$ conditioned on the initial state $s_0$. Most existing methods
independently sample the subgoal at each step from a learned prior distribution~\cite{Pertsch2020LongHorizonVP} or a replay buffer~\cite{Eysenbach2019SearchOT}, which is unlikely to propose useful plans for tasks with large, combinatorial state spaces (i.e., with multiple objects).

We propose to break down $p(\hat{s}_1, ..., \hat{s}_K | s_0)$ into $p(\hat{s}_1 | s_0) \Pi_{i=1}^k p(\hat{s}_i | \hat{s}_{i - 1})$ through modeling the conditional distribution $p(s' | s)$ of the reachable next subgoal $s'$. By utilizing temporal compositionality, the conditional subgoal generation paradigm improves generalization and enables generation of sequences of arbitrary lengths.

We use a conditional variational encoder (CVAE)~\cite{sohn2015cvae} to capture the distribution of reachable goals $p(s' | s)$. In the CVAE, we define the decoder as $g(s, z)$ and the encoder as $q(z | s, s')$, where $z$ is the learned latent representation of the transitions and it is sampled from a prior probability $p(z)$. To propose a sequence of subgoals, we use $g(s, z)$ as the conditional subgoal generator. Conditioned on the initial state $s_0$, the first subgoal $\hat{s}_1$ can be generated as $\hat{s}_1 = g(s_0, z_1)$ given the sampled $z_1 \sim p(z)$. Then the $i_\text{th}$ subgoal can be recursively generated by sampling $z_i \sim p(z)$ and computing $\hat{s}_i = g(\hat{s}_{i - 1}, z_i)$ given the previous subgoal $\hat{s}_{i - 1}$. In this way, we could sample a sequence of i.i.d. latent representations $z_1, ..., z_K$ and recursively generate $\hat{s}_1, ..., \hat{s}_K$ conditioned on the initial state $s_0$ using the conditional subgoal generator.

The CVAE is trained to minimize the evidence lower bound (ELBO)~\cite{kingma2014vae} of $p(s' | s)$ given the offline dataset $\mathcal{D}$. During training, we sample transitions $(s_t, s_{\tau})$ from the offline dataset to form the minibatches, where $\tau = t + \Delta t$ is a future step that is $\Delta t$ steps ahead. Instead of using a fixed $\Delta t$, we sample $\Delta t$ from a range for each transition to provide richer data. To encourage the trained model to be robust to compounding errors, we sample sequences composed of multiple states and use the subgoal reconstructed at the previous step as the context in the next step. Therefore, the objective for training the conditional subgoal generator is:
\begin{equation}
    \mathbb{E}_{q(z | s_t, s_{\tau})}||s_{\tau} - g(s_t, z)||^2 + \ensuremath{D_{KL}[q(z | s_t, s_{\tau}) || p(z)]}
    \label{eqn:elbo}
\end{equation}
where $\ensuremath{D_{KL}[\cdot || \cdot]}$ indicates the KL-Divergence.

% ----------------------------------------------------------------------------------------
\subsection{Efficient Planning in the Latent Space}

% ----------------------------------------------------------------------------------------
\begin{algorithm}[t]
\caption{$Plan(s_0, s_g, L, K, M, N)$}
\begin{algorithmic}[1]
\Require the initial state $s_0$, the goal state $s_g$, number of subgoals $K$, number of levels $L$, multiplier $M$, number of samples $N$.

\State Sample $N$ latent action sequences $\{z_{1:K}^i\}_{i=1}^N$.
\State Recursively generate subgoals $\{\hat{s}_{1:K}^i\}_{i=1}^N$ using $g(s, z)$.
\State Select $z_{1:K}^*$ and $\hat{s}_{1:K}^*$ of the lowest cost.
\State Update $z_{1:K}^*$ and $\hat{s}_{1:K}^*$ using MPPI.

\If{L = 1}
    \State \Return $\hat{s}_{1:K}^*$
\Else
    \State Denote $\hat{s}_0^* \leftarrow s_0$.
    \State Initialize the plan $\hat{\mathcal{S}}$ as an empty list
    \For{$i = 1, ..., K$}
        \State Append $Plan(\hat{s}_{i-1}^*, \hat{s}_{i}^*, L - 1, M, M, N)$ to $\hat{\mathcal{S}}$
    \EndFor
    \State \Return $\hat{\mathcal{S}}$
\EndIf

\end{algorithmic}
\label{algo:planning}
\end{algorithm}

We build a planner that efficiently searches for sequences of subgoals in the latent space as shown in Algorithm~\ref{algo:planning}. To tackle the large search space of candidate subgoal sequences, we design a hierarchical planning algorithm that searches for subgoals in a coarse-to-fine manner and re-use the previously selected subgoals as candidates in new episodes.

The hierarchical planning is conducted at $L$ levels with different temporal resolutions $\Delta t_1$, ..., $\Delta t_L$. The temporal resolution of each level is an integral multiple of that of the previous level, \ie, $\Delta t_i = M \Delta t_{i - 1}$, where $M$ is a scaling factor and is set to 2 in our experiments. We first plan for the subgoals $\hat{s}_1^1, \hat{s}_2^1, ...$ on the first level. Then the subgoals $\hat{s}_{1:K}^l$ of finer temporal resolution are planned on each level $l$ to connect the subgoals planned on the previous level $l-1$. Specifically, given the adjacent subgoals $\hat{s}_i^{l-1}$ and $\hat{s}_{i+1}^{l-1}$ produced on the previous level, we plan for a segment of  $M$ subgoals $\hat{s}_{i * M + 1}^{l}, ..., \hat{s}_{(i+1) * M}^{l}$ on the level $l$, by treating $\hat{s}_i^{l-1}$ and $\hat{s}_{i+1}^{l-1}$ as the initial state and final goal state in Eqn.~\ref{eqn:cost_function}. The planned segments are returned to the previous level and concatenated as a more fine-grained plan. For this purpose, we train $L$ conditional subgoal generators to propose subgoals that are $\Delta t_1$, ..., $\Delta t_L$ steps away, respectively. In contrast to the prior work~\cite{Pertsch2020LongHorizonVP}, the conditional subgoal generators enable us to plan for unseen goals that are beyond the temporal horizon of the demonstrations in the offline dataset by exploiting the compositional structure of the demonstrations. By recursively generating the subgoals across time at each level, we only need to enforce that the temporal resolution of the top level $\Delta t_L$ is smaller than the time horizon of each trajectory, since the conditional subgoal generator $f^{1}(s, z)$ needs to be trained on trajectories at least $\Delta t_L + 1$ steps in length. 

We maintain a latent plan buffer for each level to further facilitate the planning with the conditional subgoal generator. After each episode, the selected latent representations on each level are appended to the corresponding latent plan buffer. In each target task, the subgoals are supposed to have the same semantic meaning. In spite of the variations of the initial and goals state in each episode, the optimal plans in the latent space can often be similar to each other. Therefore, we sample half of the latent representations from the prior distribution $p(z)$ and the other half from the latent plan buffer among the initial samples to enhance the chance of finding a close initial guess. 

We build our planner upon the model predictive path integral (MPPI)~\cite{Gandhi2021RobustMP}, which iteratively optimizes the plan through importance sampling. In each interaction, we perturb the chosen plan in the latent space with a small Gaussian noise as new candidates. 

\begin{figure}[t!]
    \centering
    \includegraphics[width=.48\textwidth]{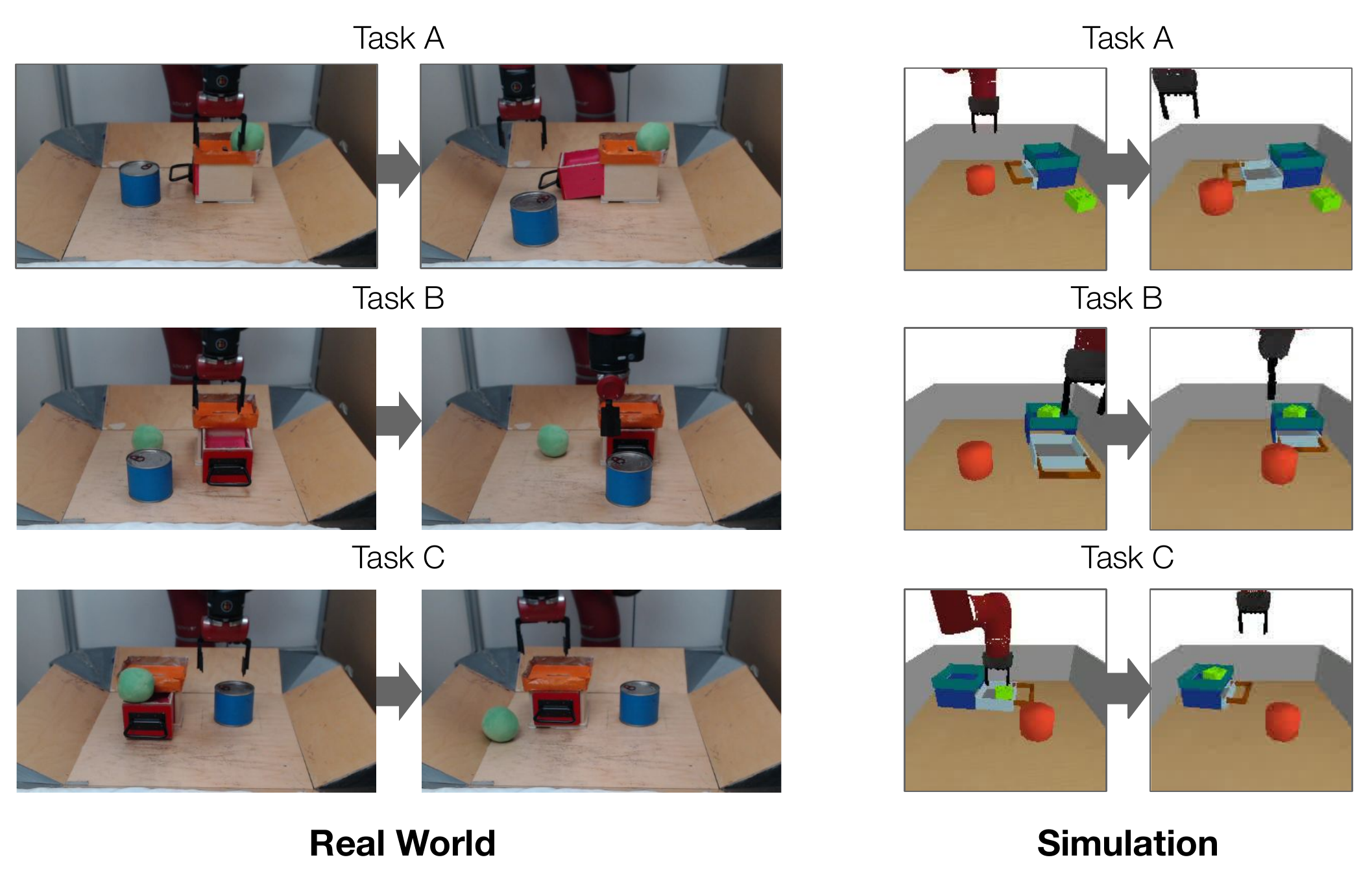}
    % \vspace{-3mm}
    \caption{\textbf{Target tasks.} Three multi-stage tasks are designed for our experiments in the simulation and the real world respectively. In each target task, the robot needs to strategically interacts with the environment (\eg, first takes out an object in the drawer then closes the drawer). The initial state and the desired goal state are shown for each task. }
    \vspace{-5mm}
    \label{fig:target_tasks}
\end{figure}

\begin{figure*}[t!]
    \centering
    \includegraphics[width=0.92\textwidth]{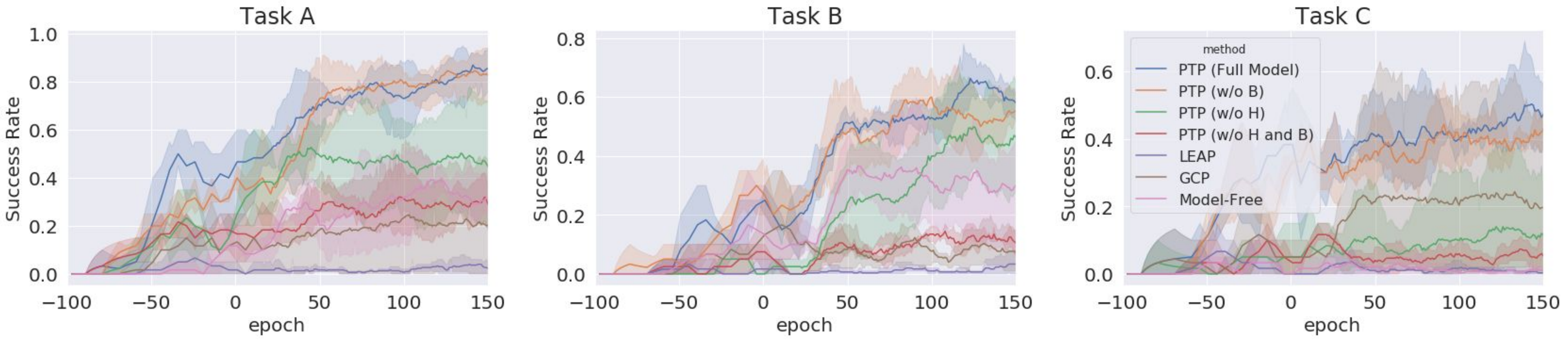}
    % \vspace{-3mm}
    \caption{
    \textbf{Quantitative comparison in simulation.} The average success rate across 3 runs is shown with the shaded region indicating the standard deviation. The negative x-axis indicates the epochs of offline pre-training and positive x-axis indicates epochs of online fine-tuning.  
    Using offline learning and planning, our method PTP is able to solve these tasks partially (at 0 epochs).
    Then with online finetuning the performance improves further.
    In contrast, prior methods have lower offline performance and do not fine-tune successfully in most cases, as they do not collect coherent online data.
    }
    \vspace{-5mm}
    \label{fig:sim_quantitative}
\end{figure*}

% ----------------------------------------------------------------------------------------
\subsection{Cost Function For Feasible Subgoals}

To provide informative guidance to the policy $\pi(a | s, s_g)$, we would like that the final goal $s_g$ can be reached at the end of the episode while encouraging the transition between each pair of subgoals to be feasible within a limited time budget. As explained in Sec.~\ref{sec:preliminaries}, the goal state is considered to be reached when the Euclidean distance between the last subgoal in the plan and the desired goal is close in the learned latent space. Inspired by \cite{Nasiriany2019PlanningWG}, the feasibility of each transition between adjacent subgoals can be measured using the goal-conditioned value function $V(s, s')$ and the log likelihood of the latent $\log p(z)$. Therefore, finding subgoals $\hat{s}_{1:K}^*$ can be formulated as the optimization problem: 
\begin{eqnarray}\label{eqn:constrained_optimization}
    \text{minimize} 
    & \quad & ||s_g - \hat{s}_K|| \\
    \text{subject to} 
    & \quad & V_{i, i+1} \geq \delta_1, \log p(z_i) \geq \delta_2, \text{for $i = 1, ..., K$} \nonumber
    % & \quad & V(\hat{s}_{i-1}, \hat{s}_{i}) \geq \delta_1, \text{for $i = 1, ..., K$} \nonumber \\
    % & \quad & \log p(z_i) \geq \delta_2, \text{for $i = 1, ..., K$} \nonumber
\end{eqnarray}
where $\delta_1$ and $\delta_2$ are thresholds and we define $\hat{s}_0 = s_0$, $V_{i, i+1} = V(\hat{s}_{i-1}, \hat{s}_{i})$ for convenience. Then the cost function can be written with Lagrangian multipliers $\eta_1$ and $\eta_2$:
\begin{eqnarray}
    && c(s_0, s_g, z_{1:K}) \\ & = & ||s_g - \hat{s}_K|| - \sum_{i=1}^{K} \left( \eta_1 V_{i, i+1} + \eta_2  \log p(z_i) \right)\nonumber
    \label{eqn:cost_function}
\end{eqnarray}
% where $\eta$ is a weight that balances the two terms. 
The details of our method are explained in Sec.~\ref{sec:implementation_details}.

%% file: 5-experiments.tex
\section{Experiments}

In our experiments, we aim to answer the following questions: 1) Can PTP propose and select feasible subgoals as plans for real-world robotic manipulation tasks? 2) Can the subgoals planned by PTP facilitate online fine-tuning of the goal-conditioned policies to solve target tasks unseen in the offline dataset? 3) How does each design option affect the performance of PTP? 
% Videos of our experimental results are available on the project website: \href{https://sites.google.com/view/planning-to-practice}{sites.google.com/view/planning-to-practice}

\subsection{Experimental Setup}
\label{sec:experimental_setup}

% We construct a simulated platform to evaluate multi-step manipulation tasks using a real-time physics simulator [16]. As shown in Fig. 1, the workspace setup includes a 7-DoF Sawyer robot arm, a table surface, and a depth sensor (Kinect2) installed overhead. Up to 5 objects are randomly drawn from a subset of the YCB Dataset [17] and placed on the table. The Sawyer robot holds a short stick as the tool to interact with the objects to complete a specified task goal.

\textbf{Environment.}
As shown in Fig.~\ref{fig:target_tasks}, our experiments are conducted in a table-top manipulation environment with a Sawyer robot. At the beginning of each episode, a fixed drawer and two movable objects are randomly placed on the table. The robot can change the state of the environment by opening/closing the drawer, sliding the objects, and picking and placing objects into different destinations, \etc. At each time step, the robot receives a 48 x 48 RGB image via a Logitech C920 camera as the observation and takes a 5-dimensional continuous action to change the gripper status through position control. The action dictates the change of the coordinates along the three axes, the change of the rotation, and the status of the fingers. We use PyBullet~\cite{coumans2016pybullet} for our simulated experiments.  

\textbf{Prior data.}
The prior data consists of varied demonstrations for different primitive tasks. In each demonstrated trajectory, we randomly initialize the environment and perform primitive interactions such as opening the drawer and poking the object. These trajectories are collected using teleoperation in the real world, and a scripted policy that uses privileged information of the environment (e.g., the object pose and the status of the drawer) in simulation. The trajectories vary in length from 5 to 150 time steps, with 2,344 trajectories in the real world and 4,000 in simulation.

\textbf{Target tasks.} 
In each target task, a desired goal state is specified by a 48 x 48 RGB image (same dimension with the observation). The robot is tasked to reach the goal state by interacting with the objects on the table. Task success for our evaluation is determined based on the object positions at the end of each episode (this metric is not used for learning). As shown in Fig.~\ref{fig:target_tasks}, we design three target tasks that require multi-stage interactions with the environment to complete. These target tasks are designed with temporal dependencies between stages (\eg, the robot needs to first move away a can that blocks the drawer before opening the drawer). The transitions from the initial state to the goal state are unseen in the offline data. The episode length is 400 steps in simulation and 125 steps in the real world, which are much longer than the time horizon of the demonstrations. 

\textbf{Baselines and ablations.} 
We compare PTP with 3 baselines and 3 ablations. \textbf{Model-Free} uses a policy directly conditioned on the final goal and conducts online fine-tuning without using any subgoals. \textbf{LEAP}~\cite{Nasiriany2019PlanningWG} learns a variational auto-encoder (VAE)~\cite{kingma2014vae} to capture the prior distribution of states and plans for subgoals without conditioning on any context. \textbf{GCP}~\cite{Pertsch2020LongHorizonVP} learns a goal predictor that hierarchically generates intermediate subgoals between the initial state and the goal state. To analyze the design options in PTP, we also compare with variations of our method by removing the latent plan buffer (\textbf{PTP (w/o B)}), the hierarchical planning algorithm (\textbf{PTP (w/o H)}), and both of these two designs (\textbf{PTP (w/o H and B)}). All methods use the same neural network architecture in the goal-conditioned policy and are pre-trained on the same offline dataset.

\subsection{Implementation Details}
\label{sec:implementation_details}

Following \cite{Khazatsky2021WhatCI}, we use a vector quantized variational autoencoder (VQ-VAE)~\cite{Oord2017NeuralDR} as the state encoder, which encodes a $48 \times 48 \times 3$ image to a 720-dimensional encoding. The conditional subgoal generator is implemented with a U-Net architecture~\cite{Ronneberger2015UNetCN} and decodes the subgoal from an 8-dimensional latent representations conditioned on the encoding of the current state. In our planner, we use $L = 3$, $K = 8$, $M = 2$, $N = 1024$, and we run MPPI for 5 iteration on each level. $g$ is trained to predict subgoals that are 15, 30, and 60 steps away. Implicit Q-Learning (IQL)~\cite{kostrikov2021iql} is used as the underlying RL algorithm for offline pre-training and online fine-tuning with default hyperparameters. We use the same network architectures for the policy and the value functions from ~\cite{Khazatsky2021WhatCI} for simulation experiments. For real-world experiments, we use a convolutional neural network instead. We use Adam optimizer with a learning rate of $3 \cdot 10^{-4}$ and a batch size of 256. During training, we relabel the goal with future hindsight experience replay~\cite{Andrychowicz2017HindsightER} with $70\%$ probability. We use $\epsilon=2$ for the reward function defined in Sec.~\ref{sec:preliminaries}, $\eta_1 = 0.001, \eta_2 = 0.01$ in Eqn.~\ref{eqn:cost_function}. 

\begin{figure}[t!]
    \centering
    \includegraphics[width=.48\textwidth]{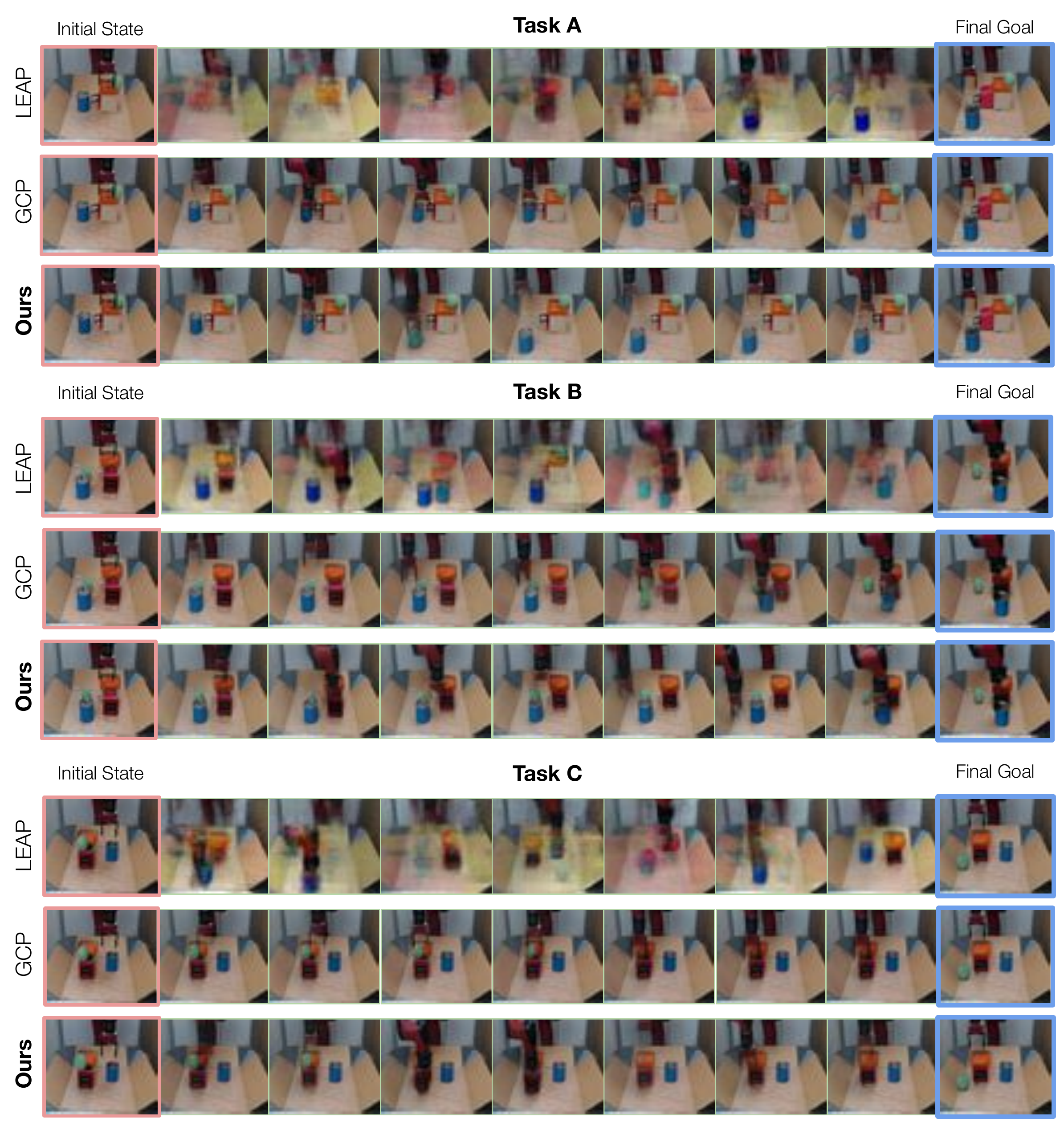}
    \caption{\textbf{Planned subgoal sequences.} Each row shows the sequence of subgoals produced by each method. The initial state and the final goal are shown at the two ends. 
    }
    \vspace{-5mm}
    \label{fig:real_qualitative}
\end{figure}

\subsection{Quantitative Comparisons}

We evaluate PTP and baselines on three unseen target tasks. We use simulated versions of these tasks for comparisons and ablations, and real-world tasks, where all pretraining and finetuning uses only real-world data, to evaluate the practical effectiveness of the method.

\textbf{Simulation}. We first pre-train the goal-conditioned policy on the offline dataset for 100 epochs and the run online fine-tuning for the target task for 150 epochs. Each epoch takes 2,000 simulation steps (only during fine-tuning) and 2,000 training iterations. We run online fine-tuning using each method with 3 different random seeds. After each epoch, we test the policy in the target task for 5 episodes. We report the average success rate across 3 runs in Fig.~\ref{fig:sim_quantitative} where the negative x-axis indicates the offline pre-training epochs and positive x-axis indicates the online fine-tuning epochs.

As shown in Fig.~\ref{fig:sim_quantitative}, our full model consistently outperforms baselines with a large performance gap. The generated subgoals not only enable the pre-trained policy to achieve higher success rate by breaking down the hard problems into easier pieces, but also introduces larger performance improvements during online fine-tuning. After fine-tuning for 150 epochs, the policy achieves the success rates of 84.9\%, 59.9\%, 49.3\% in the three target tasks respectively. Compared to the policy pre-trained on the offline dataset, the performance is significantly improved (+31.6\%, +37.8\%, and +13.8\%). When directly using the final goal or subgoals generated by baseline methods, the policy's performance plateaus at around 0.0\% to 30.0\% and does not improve much during online fine-tuning. 

We found that the hierarchical planner and the latent plan buffer are crucial for PTP's performance. Without these two design options, the planner often suffers from the large search space of possible subgoal sequences and the resultant success rates decrease. The latent plan buffer significantly improves the performance of non-hierarchical PTP while it has a minor effect on hierarchical PTP.   

\textbf{Real-world evaluation.} We pre-train the policy for 200 epochs and fine-tune it for 10 epochs. In each epoch, we run 10,000 training iterations and collect 1,000 steps in the real world. We train on three target tasks which are shown in Figure~\ref{fig:target_tasks}, and report the success rate of the goal-conditioned policy before and after online fine-tuning in Table~\ref{fig:real_quantitative}. Planning enables the robot to succeed partially with just the offline initialized policy, achieving success rates of $12.5\%, 75.0\%$ and $25.0\%$ on the three tasks. (When the offline policy is conditioned on only the final goal image without planning, the success rate is $0\%$.) Then in each task, we fine-tune to a significantly higher success rate.

Qualitatively, at the beginning of fine-tuning, the robot often fails, deviating from the planned subgoals or colliding with the environment.
With the planned subgoals, the original long-horizon task is broken down to short snippets that are easier to complete.
Even if a subgoal is not reached successfully at first, the data is useful to collect additional experience and fine-tune the policy.
After fine-tuning for 4-5 epochs, we already observe that the robot's performance reaching subgoals during training time significantly improves, collecting even more coherent and useful data.
After 10 epochs, we achieve success rates of $62.5\%, 100.0\%$ and $50.0\%$.
In comparison, GCP cannot provide useful guidance to the policy when the generated goals are noisy.

\vspace{+2mm}
\begin{table}[t!]
    \normalsize
    \centering
    \caption{The real-world success rates before and after online fine-tuning. The tasks are described in Sec.~\ref{sec:experimental_setup}.}
    \begin{tabular}{ c|c|c }
        \centering
        Task & \begin{tabular}{@{}c@{}}PTP (Ours) \\ Offline $\rightarrow$ \text{Online} \end{tabular} & \begin{tabular}{@{}c@{}} GCP \\ Offline $\rightarrow$ \text{Online} \end{tabular} \\
        \hline
        Task A & $12.5\% \rightarrow \textbf{62.5\%}$ & $12.5\% \rightarrow 0.0\%$ \\
        Task B & $75.0\% \rightarrow \textbf{100.0\%}$ & $50.0\% \rightarrow 75.0\%$ \\
        Task C & $25.0\% \rightarrow \textbf{50.0\%}$ & $25.0\% \rightarrow 12.5\%$ \\
    \end{tabular}
    \label{fig:real_quantitative}
    \vspace{-6mm}
\end{table}
% \end{figure}

\subsection{Generated Subgoals}

In Fig.~\ref{fig:real_qualitative}, we present qualitative results of the generated subgoals for each task in the real world. Each row shows a sequence of generated subgoals produced by the planner in each method. In all the three target tasks, PTP successfully plans for a sequence of subgoals that can lead to the desired final goal. The transition between adjacent subgoals are feasible within a short period of time. By comparison, both of the baseline methods fail to generate reasonable plans. Without conditioning on the current state, LEAP~\cite{Nasiriany2019PlanningWG} can hardly produce any realistic images of the environment. Most of the generated subgoals are highly noisy images with duplicated robot arms and objects. The quality of the subgoals produced by GCP~\cite{Pertsch2020LongHorizonVP} is higher than that of LEAP but still much worse than ours. GCP cannot generalize well for the initial state and the goal state that are out of the distribution of the offline dataset, which contains only short snippets of demonstrations.

%% file: 6-conclusion.tex
\section{Conclusion and Discussion}

We presented PTP, a method for real-world learning of temporally extended skills by utilizing planning and fine-tuning to stitch together skills from prior data.
First, planning is used to convert a long-horizon task into achievable subgoals for a lower level goal-conditioned policy trained from prior data.
Then, the goal-conditioned policy is further fine-tuned with active online interaction, mitigating the distribution shift between the offline data and actual states seen during rollouts.
This procedure allows robots to extend their capabilities autonomously, composing previously seen data into more complicated and useful skills.

\section*{Acknowledgement} 

Our research is supported by the Office of Naval Research, ARL DCIST CRA W911NF-17-2-0181, ARO W911NF-21-1-0097, and AFOSR FA9550-22-1-0273. We thank Dhruv Shah and Ilya Kostrikov for their constructive feedback.